\documentclass{article}
\usepackage[a4paper, total={6in, 8in}]{geometry}
\usepackage[utf8]{inputenc}
\usepackage{amssymb}
\usepackage{amsmath}
\usepackage{mathtools}
\usepackage{amsthm}
\usepackage{color}
\usepackage{dsfont}
\usepackage[square,numbers]{natbib}
\usepackage{hyperref}
\hypersetup{
    colorlinks=true,
    linkcolor=blue,
    filecolor=magenta,      
    urlcolor=cyan,
    }
\usepackage{adjustbox}

\newtheorem{theorem}{Theorem}[section]
\newtheorem{lemma}[theorem]{Lemma}

\makeatletter
\newcommand{\distas}[1]{\mathbin{\overset{#1}{\kern\z@\sim}}}%
\newsavebox{\mybox}\newsavebox{\mysim}
\newcommand{\distras}[1]{%
  \savebox{\mybox}{\hbox{\kern3pt$\scriptstyle#1$\kern3pt}}%
  \savebox{\mysim}{\hbox{$\sim$}}%
  \mathbin{\overset{#1}{\kern\z@\resizebox{\wd\mybox}{\ht\mysim}{$\sim$}}}%
}

 \DeclareMathOperator{\trace}{Trace}
 \DeclareMathOperator{\sg}{SG}
 \DeclareMathOperator{\reg}{reg}

\title{Deep Unsupervised Feature Selection by Discarding Nuisance and Correlated Features}
\author{Uri Shaham\thanks{equal contribution}, Ofir Lindenbaum\footnotemark[1], Jonathan Svirsky, Yuval Kluger}
\date{October 2021}

\begin{document}

\maketitle

\begin{abstract}
Modern datasets often contain large subsets of correlated features and nuisance features, which are not or loosely related to the main underlying structures of the data. 
Nuisance features can be identified using the Laplacian score criterion, which evaluates the importance of a given feature via its consistency with the Graph Laplacians' leading eigenvectors.
We demonstrate that in the presence of large numbers of nuisance features, the Laplacian must be computed on the subset of selected features rather than on the complete feature set.
To do this, we propose a fully differentiable approach for unsupervised feature selection, utilizing the Laplacian score criterion to avoid the selection of nuisance features.
We employ an autoencoder architecture to cope with correlated features, trained to reconstruct the data from the subset of selected features.
Building on the recently proposed concrete layer that allows controlling for the number of selected features via architectural design, simplifying the optimization process.
Experimenting on several real-world datasets, we demonstrate that our proposed approach outperforms similar approaches designed to avoid only correlated or nuisance features, but not both. Several state-of-the-art clustering results are reported.
Our code is publicly available at~\url{https://github.com/jsvir/lscae}.

\end{abstract}

\section{Introduction}\label{sec:intro}
Feature Selection is an important area of machine learning. Reducing data dimensionality may be appealing for numerous reasons, for example, reduction of the computational burden, overfitting prevention, and simplification of downstream tasks and analyses, to name a few.
In the supervised learning regime, it is straightforward to evaluate the quality of different subsets of input features simply by measuring the generalization performance of a model given the selected features as inputs.
Evaluation of feature selection in the unsupervised regime, on the other hand, is a more involved task, as there is no natural (single) criterion for evaluation of the quality of different subsets of selected features.

We identify two dominant feature evaluation criteria in the unsupervised feature selection literature. 
The first criterion is the consistency of a feature with the main underlying structures of the data.
It is common to associate these main underlying structures with the leading eigenvectors of the graph Laplacian matrix of the data. 
It is well known that when the data is clusterable, the cluster structure can be recognized in the subspace of Laplacian's leading eigenvectors~\cite{von2007tutorial}.
In addition, when the data has a low-dimensional structure (e.g., lies on a manifold), the diffusion distance~\cite{coifman2006diffusion}, which describes the similarity between data points, is governed by the large eigenvalues of the Laplacian.

Thus, a common means to measure the consistency of a feature with the main underlying data structures is by the inner product of the given feature with the leading eigenvectors of the graph Laplacian of the data~\cite{he2006laplacian, cai2010unsupervised,zhao2007spectral, wang2015embedded}, which is known as the ``\textit{Laplacian score}'' criterion. 
Thus, the Laplacian score criterion favors features that have significant components in the subspace of the leading Laplacian eigenvectors.
Selection of features that respect the multi-cluster structure was also approached via alternative measures; see, for example~\cite{yang2011}.

The second feature evaluation criterion is the amount of information a feature contains on other features in the dataset. 
The logic behind this criterion is that a subset of representative features is helpful for selection if it carries a sufficient amount of information to represent the complete feature set, despite being sparse.
Thus, a common means to achieve this goal is to search for a subset of features from which the full feature set can be approximately reconstructed.
Such an approach can be found in~\cite{zhu2015unsupervised, balin2019concrete, AE1, AE2}.
It is easy to see how this criterion can yield a highly sparse set of selected pixels on the MNIST handwritten dataset, for example, while capturing sufficient information to reconstruct all pixels, as neighboring pixels in this dataset tend to be highly correlated.

Each of the two above criteria implicitly defines an inductive bias, which specifies the characteristics of features that should be selected.
The first criterion favors features which correspond to the main structures (e.g., cluster structures or ``slow-varying'' ones) in the data, hence tends to discard features which do not manifest this structure; such features are often called ``\textit{nuisance features}'', as opposed to ``informative features'', which do carry information on the data underlying structures.
The second criterion favors features that are correlated with a large number of other features and can thus be used to reconstruct the complete feature set approximately. 

Realistically, however, modern datasets contain both types of features (i.e., nuisance ones and correlated ones), making any feature selection method that is designed to handle only one type of features sub-optimal
\footnote{
In this sense, a comparison between methods corresponding to different criteria, often found in the unsupervised feature selection literature, may be comparing apples and oranges, as the differences in performance can depend more on the types of features which exist in the dataset on which the comparison is made, and less on algorithmic matters.}.
Therefore combining the two evaluation criteria can arguably be more suited for general unsupervised feature selection purposes. 
Despite that, to the best of the authors' knowledge, most of the existing unsupervised feature selection methods are not designed to handle both correlated and nuisance features. 

Moreover, while the Laplacian score is a widely used tool for evaluating feature importance, it is commonly computed based on a Laplacian that relies on the complete feature set.
However, in the presence of a large number of nuisance features, the Laplacian gets corrupted in the sense that its leading eigenvectors no longer correspond to the manifold or cluster structure of the data.
We demonstrate this problematic aspect of the Laplacian score. We argue that the Laplacian should be computed on the subset of selected features to avoid it rather than on the complete feature set.
While this makes the selection process more involved, it can be stated as a differentiable cost function; hence it can be solved using tools commonly used in deep learning.

Being a convenient framework for optimizing differentiable objective functions, deep learning algorithms are widely used for many traditional machine learning tasks. 
This is the case also in unsupervised feature selection, where several recently proposed methods are implemented as deep learning algorithms. For example~,\cite{balin2019concrete, AE1, AE2} are autoencoder-based methods, which aim to reconstruct the data from a small subset of selected features, hence correspond to the second criterion. \cite{lindenbaum2020let} is based on stochastic gates and aims to find a small subset of selected features that maximize the Laplacian score, hence corresponds to the first criterion.

In this work, we follow this direction and propose a deep learning method for unsupervised feature selection, which aims to correspond to the two above criteria by discarding both nuisance and correlated features in a differential fashion.
Our method builds on the recently proposed Concrete Autoencoder (CAE,~\cite{balin2019concrete}), augmenting its objective function using a differential Laplacian score term.
CAE is equipped with a Concrete layer, which controls the number of selected features using an elegant architectural design; we utilize this mechanism to tackle the problematic aspects of computing the Laplacian score for the complete feature set.

To highlight the utility of the proposed approach, we first simulate a scenario in which it stands out compared to related methods that handle either nuisance or correlated features, but not both, in the sense that it allows better recognition of the data true cluster structure.
We then report experimental results on ten real-world datasets, demonstrating that the proposed approach can lead to state-of-the-art downstream clustering tasks.

Our contributions are four-fold:
(i) We demonstrate that a large number of nuisance features corrupts the Laplacian, making the Laplacian score a sub-optimal measure for feature quality; we also provide analytical arguments supporting this claim.
(ii) We propose an autoencoder-based approach for unsupervised feature selection, designed to handle both correlated and nuisance features.
(iii) We experimentally demonstrate the advantage of the proposed approach over methods that consider either correlated features or nuisance ones, but not both, on several real-world unsupervised feature selection benchmarks, reporting state-of-the-art performance.
(iv) We provide a user-friendly Python implementation of the proposed approach for general use.

This work builds on an earlier work of ours~\citet{lindenbaum2020let}, which contains only the Laplacian score objective and utilizes stochastic gates for the selection mechanism. In particular Section~\ref{sec:motivation} appears in similar form therein. Yet, the proposed methodology in the current manuscript differs significantly from the one in~\citep{lindenbaum2020let}.

The remainder of this manuscript is organized as follows.
In Section~\ref{sec:preliminaries} we review preliminary materials. 
Our proposed approach is motivated in Section~\ref{sec:motivation} and described in Section~\ref{sec:proposed}.
Experimental results are provided in Section~\ref{sec:experiments}.
Section~\ref{sec:conclusion} briefly concludes the manuscript.


\section{Preliminaries}\label{sec:preliminaries}

Consider a data matrix $X \in \mathbb{R}^{n\times d}$ with $d$-dimensional observations $x_1,\ldots, x_n$. We refer to the columns of $X$ as features $f_1,\ldots, f_d$, and we assume that the features are centered and normalized, i.e., for each $i$, $1^Tf_i = 0$ and $\|f_i\|_2^2=1$.

\subsection{Laplacian score}\label{sec:ls}

Given $n$ data points, a kernel matrix is a $n \times n$ matrix $K$, whose $(i,j)$ entry quantifies the similarity between $x_i$ and $x_j$. For example, in many applications, such matrix is constructed using a Gaussian kernel 
$$
K_{i,j} = \exp\left({-\frac{\| x_i-x_j\|^2}{2\sigma^2}}\right),
$$
where $\sigma$ is a user-defined parameter that determines the sensitivity of the kernel\footnote{A common practice to choose $\sigma$ is to set it to the maximal Euclidean distance from any point to its nearest neighbor, and many other practices exist as well.}.

Given a kernel matrix $K$, the unnormalized graph Laplacian $L_\text{un}$ is defined via $L_\text{un} = D - K$, where $D$ is a diagonal matrix of row sums of $K$.
It is common to interpret the Laplacian eigenvalues as frequencies so that eigenvectors corresponding to larger eigenvalues oscillate faster \cite{coifman2006diffusion}.
Assuming that important underlying patterns of the data (e.g., cluster structure) are slowly varying, the eigenvectors corresponding to the smallest eigenvalues of $L_\text{un}$ express the main structures of the data. This fact is used as a basis for various manifold learning and dimensionality reduction techniques.

For example, when the Laplacian represents clusterable data with $m$ distinct components, the leading $m$ eigenvectors provide a complete specification of the cluster allocation of any point; this insight led to the celebrated spectral clustering method\citep{von2007tutorial}. 
Since the leading eigenvectors of the Laplacian describe the important structures of the data, it makes sense to evaluate a feature by how much it respects this structure. 
This idea lies at the core of the Laplacian score method, proposed by~\cite{he2006laplacian}, where each feature $f$ is assigned the score 
$$\text{score}(f) = f^TL_\text{un}f = \sum_{i=1}^n \lambda_i \langle u_i, f\rangle ^2, $$
where $L_\text{un} = \sum_{i=1}^n \lambda_i u_iu_i^T$ is the eigendecomposition of $L_\text{un}$.
Therefore, the smaller the score a feature $f$ is assigned, the more significant is the component of $f$ in the subspace of the leading eigenvectors of $L_\text{un}$, implying that $f$ is more consistent with the main structures of the data, making it an essential feature for selection.

Similar behavior of the eigenvectors of the unnormalized Laplacian exists for the diffusion Laplacian $L_\text{diff} = D^{-1}K$ which expresses the transition probabilities between any pair of points, except that for $L_\text{diff}$ the eigenvectors corresponding to largest eigenvalues are the ones that express the main structures in the data.
To use the same terminology for both Laplacians, we use the term ``leading'' to refer to the eigenvectors corresponding to the smallest eigenvalues of unnormalized Laplacian and the eigenvectors corresponding to the largest eigenvectors in case of the diffusion Laplacian.

\subsection{Concrete Layer}\label{sec:cl}

The Concrete distribution~\cite{maddison2016concrete} is a continuous relaxation of discrete random variables, which allows differentiation through a sampling procedure, in a similar fashion to the reparametrization trick for continuous random variables~\cite{kingma2013auto}. This opens a wide range of deep learning applications which incorporate discrete random variables into the training procedure.

More specifically, a sample $z$ of a categorical random variable with probabilities $(\pi_1,\ldots,\pi_d)$ can be obtained via the Gumble-max trick~\cite{gumbel1954statistical}
\begin{equation}
    z = \arg\max_i(g_i + \log \pi_i), \label{eq:gm}
\end{equation}
where $g_1,\ldots, g_d$ are iid samples from a $\text{Gumble}(0,1)$ distribution.
Since the softmax function is a continuous approximation of the $\arg\max()$ function, equation~\eqref{eq:gm} can be relaxed into a continuous approximation~\cite{jang2017categorical} via
\begin{equation}
    z_i =  \frac{\exp((g_i + \log \pi_i)\slash \tau)}{\sum_{j=1}^d\exp((g_j + \log \pi_j)\slash \tau)}, \label{eq:gs}
\end{equation}
where $\tau$ is a temperature parameter, governing the extent to which the softmax vector is peaked.

In a concrete layer, each unit approximates the sampling of a single entry from its input vector. The approximation is performed by a dot product of the Gumble-Softmax vector~\eqref{eq:gs} with the input.
The temperature $\tau$ is typically annealed throughout training, starting from a high value (for which the resulting softmax vector is flat) and ending in a value near zero, for which the softmax vector is close to being one-hot.  In the latter case, the dot product approximates sampling from a categorical distribution.
The probabilities $\pi_i,\; i=1,\ldots, d$ of each unit are learnable differentiable parameters, which are trained via backpropagation.
A concrete layer of size $k$ is therefore parametrized by a $k \times d$ parameter $\Pi$, where $\Pi_{i,1},\ldots \Pi_{i,d}$ are the categorical probabilities of the $i$th concrete unit. This results in an $k\times d$ feature selection weight matrix $Z$, obtained via equation~\eqref{eq:gs}. 


\subsection{Concrete Autoencoder}\label{sec:cae}

Concrete Autoencoder (CAE,~\cite{balin2019concrete}) is a state-of-the-art deep unsupervised feature selection method, based on a standard autoencoder, having a concrete layer as its first layer.
The size of the concrete layer determines the desired number of selected features.
The autoencoder is trained by minimizing reconstruction loss using gradient based optimization.
As explained in the previous section, after the temperature annealing process, each unit in the concrete layer approximates sampling of a single input feature from a categorical distribution, which amounts to selecting a single feature.

The reconstruction error loss makes CAE select features from which the full feature set can be reconstructed. Hence CAE is designed for scenarios where there are subsets of correlated features, where each such subset can be reconstructed from a single or a few representative features. 

The concrete layer of CAE provides an elegant, architectural-based control for the number of selected features, which simplifies the training process, as the loss function can contain only the reconstruction error term. 
This alleviates the need for a regularization term to encourage sparsity of the selected subset, as in the core of several other deep learning approaches for feature selection, e.g.,~\cite{lindenbaum2020let}.


\section{Motivation}\label{sec:motivation}

This section motivates the need to compute the graph Laplacian on the subset of selected features rather than on the complete feature set when the data contains many nuisance features. 
To do so, we first present a diffusion perspective and demonstrate the change in the Laplacian eigenvalues and eigenvectors empirically as the number of nuisance features grows. 
We then consider a two-cluster case study and show analytically how the number of nuisance dimensions affects the ability to recover the main underlying structure of the data.


\subsection{A Diffusion Perspective}\label{sec:diff}
\begin{figure}[h!]
  \centering
    \includegraphics[height=.25\textwidth]{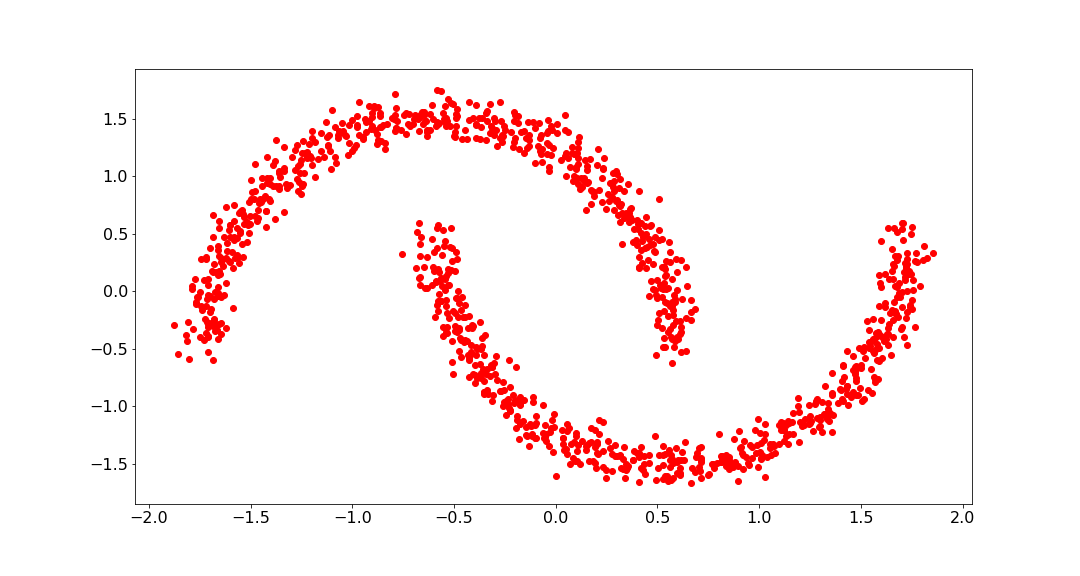}  
    \includegraphics[height=.25\textwidth]{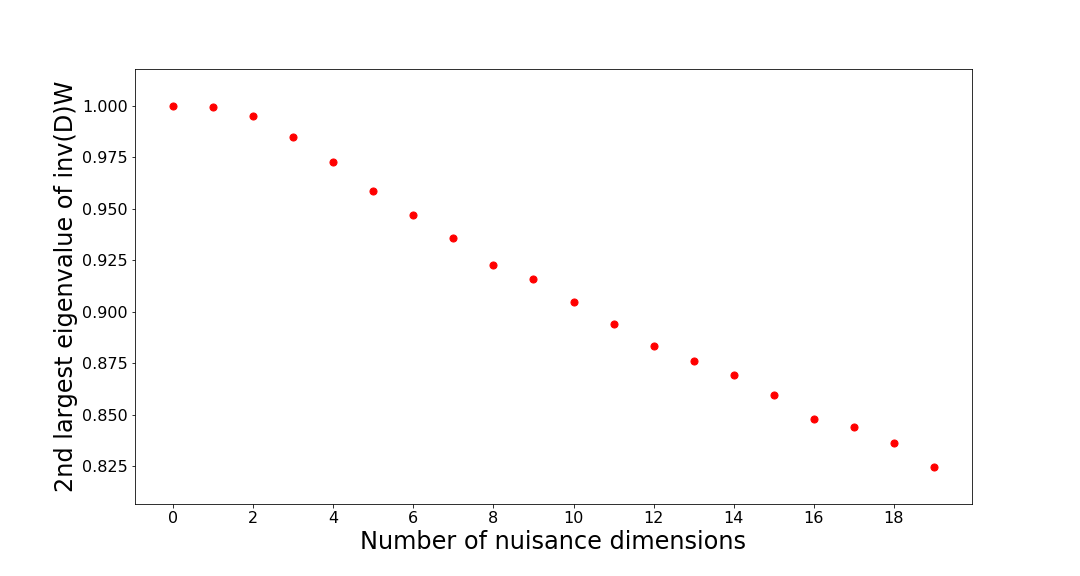}
    \includegraphics[height=.25\textwidth]{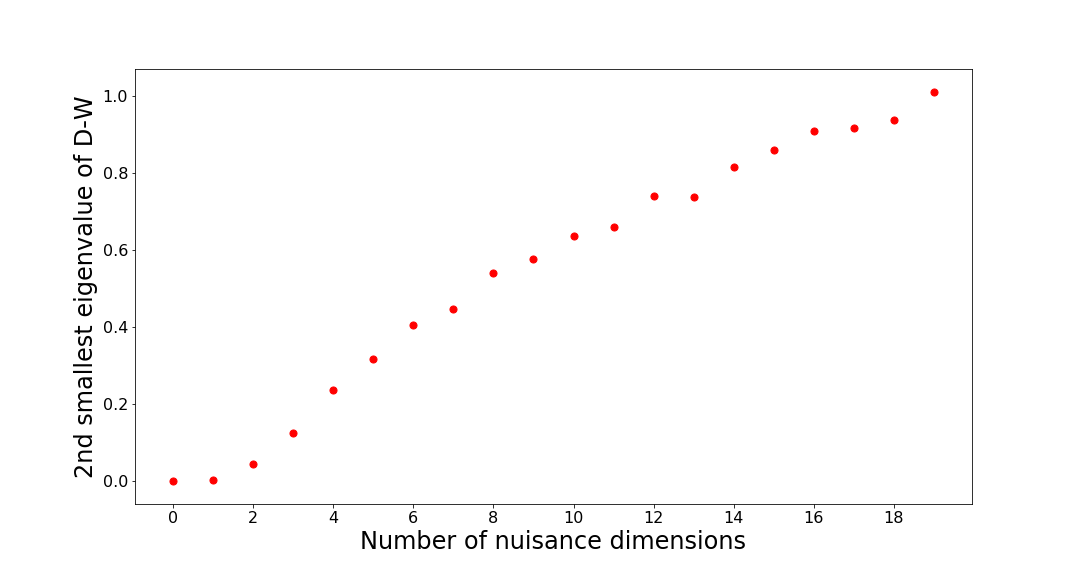}
    \includegraphics[height=.25\textwidth]{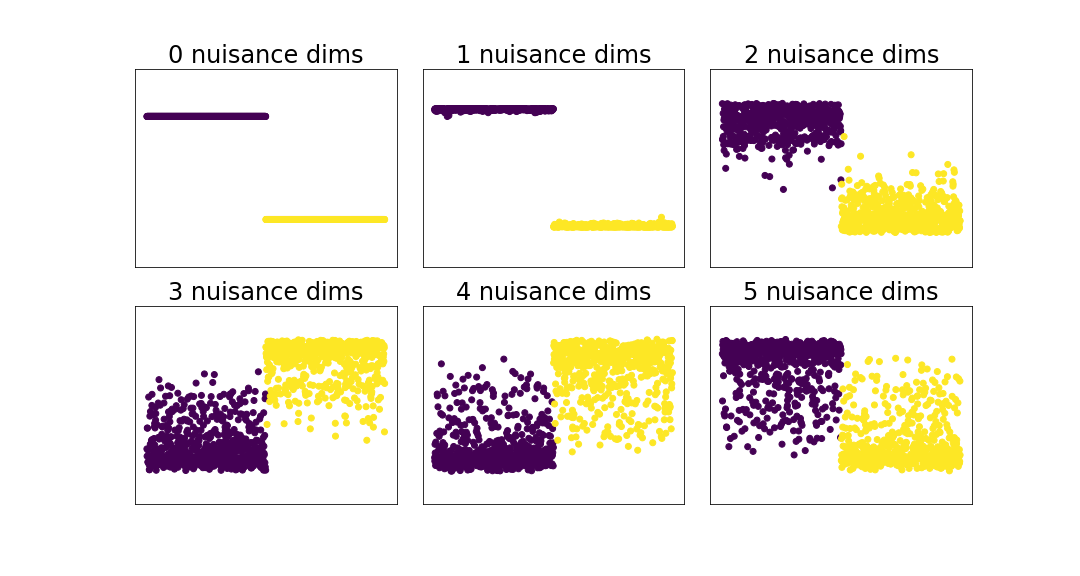}
    \includegraphics[height=.25\textwidth]{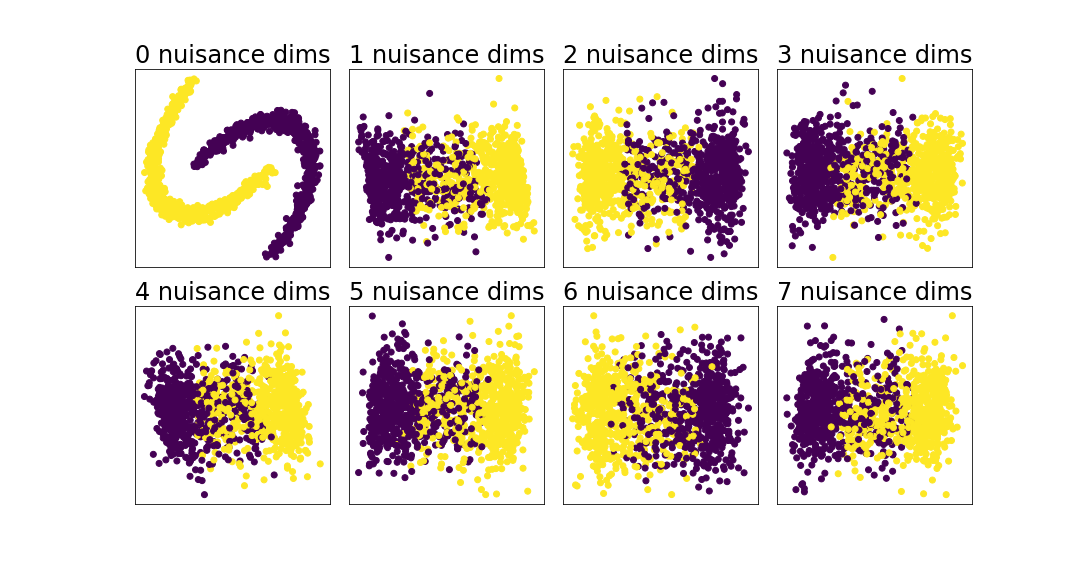}
    \includegraphics[height=.25\textwidth]{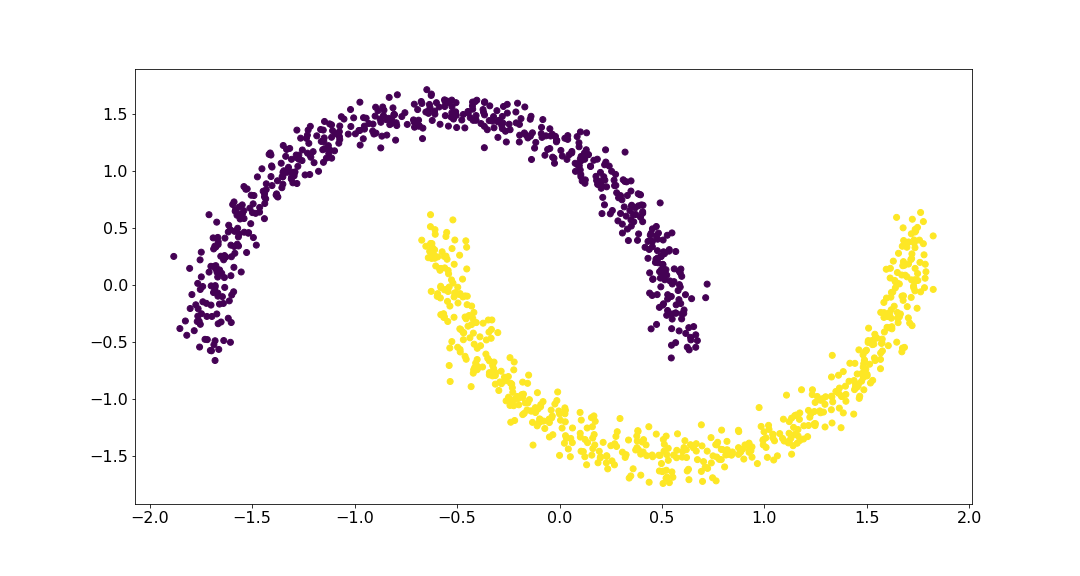}
    \caption{Top left: the Two moons dataset. Top right: the second largest eigenvalue of the diffusion Laplacian decreases as the number of nuisance dimensions grows. 
    Middle left: similarly, the algebraic connectivity of the graph increases with more nuisance dimensions. Middle right: with more nuisance dimensions, the second leading eigenvector of the graph Laplacian is no longer a clear indicator of the cluster assignments. In each subplot, the vertical position corresponds to the entry in the second leading Laplacian eigenvector; the horizontal position corresponds to the cluster assignment. Bottom left: Projection of the data on the first two principal directions. PCA cannot recover the underlying cluster structure, as the directions of maximal variance correspond to noise. Bottom right: our proposed approach identifies the informative dimensions, which enables downstream analysis, e.g., using SpectralNet~\cite{shaham2018spectralnet}.} 
    \label{fig:twomoons}
\end{figure}

Consider the simple 2-dimensional dataset, known as ``Two moons'', shown in the top left panel of Figure~\ref{fig:twomoons}, which contains two (nonconvex) separate clusters.
We extend this dataset by adding $k$ nuisance dimensions, each of which is a sample of iid $\text{unif}(0,1)$ entries.
As the number $k$ of nuisance dimensions grows, the clear cluster structure is obscured, as the amount of noise dominates the signal. Consequently, recognizing the actual underlying two-cluster structure becomes challenging and is likely to fail.

From a diffusion perspective, data are considered clusterable if a random walk that starts inside a cluster takes a long time to exit the cluster. 
The cluster exit times are manifested by the leading (i.e., largest) eigenvalues of the diffusion Laplacian $L_\text{diff} = D^{-1}W$ (for example, when the data contains $m$ completely separate clusters, and a random walk can never leave the cluster it begins at, the top $m$ eigenvalues of $L_\text{diff}$ are all equal to 1).
Each added nuisance dimension increases the variability inside any cluster and increases the distances between any point and its true nearest neighbors (that is, ones which belong to the same ``moon'' in the two-moons example). 
At the same time, the added noise is likely to create spurious similarities between points, regardless of the actual cluster they belong to.
Altogether, this shortens the cluster exit times, which is manifested by the decrease of the second largest eigenvalue of $L_\text{diff}$, as is shown in the top right panel of Figure~\ref{fig:twomoons}.
A similar behavior occurs by looking at the second smallest eigenvalue of the unnormalized Laplacian $L_\text{rw} = D - W$, known as the ``algebraic connectivity'' or Fiedler number, which increases with the number of nuisance dimensions, implying that these dimensions make the graph more connected, as is shown in Figure~\ref{fig:twomoons} as well (middle left panel).
The fact that the graph becomes more connected is also manifested by the second leading eigenvector of the Laplacian, which becomes less indicative of the correct cluster assignment as the number of nuisance dimensions grows (middle right panel).
As a result of the graph becoming more connected, attempts to recover the actual cluster structure are more likely to fail as the number $k$ of nuisance dimensions grows. 
One may argue that this can be avoided by using a dimensionality reduction technique, like PCA, to capture the signal (i.e., the correct cluster structure) and remove much noise. However, as can be seen in the bottom left panel of Figure~\ref{fig:twomoons}, using PCA in this example, unfortunately, does not capture the actual underlying structure, as the directions of maximal variance correspond to noise and not to the correct cluster structure.
Applying our proposed approach to the above dataset, the two informative dimensions are selected, and the nuisance dimensions are discarded. As a result, downstream algorithms like SpectralNet~\cite{shaham2018spectralnet} can correctly identify the cluster structure (bottom right panel).

To complement this empirical analysis, in the next section, we consider a simple two-cluster case. We analytically derive the connection between the number of nuisance dimensions and one's ability to recover the cluster structure of the data.

\subsection{Case Study Analysis}
To observe the effect of nuisance dimensions, in this section, we consider a simple example where all of the noise in the data arises from such dimensions. Specifically, consider a dataset that includes $2n$ datapoints in $\mathbb{R}$, where $n$ of which are at $0 \in \mathbb{R}$ and the remaining ones are at $r >0$, i.e., each cluster is concentrated at a specific point. Next, we add $d$ nuisance dimensions to the data so that samples lie in $\mathbb{R}^{d+1}$. The value for each data point in each nuisance dimension is sampled independently from $N(0, 0.5 ^2)$.

Suppose we construct the graph Laplacian by connecting each point to its nearest neighbors.
We would now investigate the conditions under which the neighbors of each point belong to the correct cluster. 
Consider points $x,y$ belonging to the same cluster. Then $(x-y) = (0, u_1,\ldots, u_d)$ where $u_i\distas{\text{iid}} N(0, 1)$, and therefore $\|x-y\|^2 \sim \chi^2_d$.
Similarly, if $x, y$ belong to different clusters, then  $\|x-y\|^2 \sim r^2 + \chi^2_d$.
Now, to find conditions for $n$ and $d$ under which with high probability the neighbors of each point belong to the same cluster, we can utilize Chi-square measure-concentration bounds~\cite{laurent2000adaptive}.
\begin{lemma}[\cite{laurent2000adaptive} P.1325]
Let $X\sim\chi^2_d$. Then 
\begin{enumerate}
    \item $\mathbb{P}(X-d \ge 2\sqrt{d\gamma}+2\gamma) \le \exp(-\gamma)$.
    \item $\mathbb{P}(d-X \ge 2\sqrt{d\gamma}) \le \exp(-\gamma)$.
\end{enumerate}
\label{lemma:massart}
\end{lemma}

Given sufficiently small $\gamma > 0$ we can divide the segment $[d, d + r^2]$ to two disjoint segments of lengths $2\sqrt{d\gamma}+2\gamma$ and $2\sqrt{d\gamma}$ (and solve for $d$ in order to have the total length $r^2$). 
This yields
\begin{equation}
    \sqrt{d} = \frac{r^2-2\gamma}{4\sqrt{\gamma}}.\label{eq:d}
\end{equation}
The nearest neighbors of each point will be from the same cluster as long as all distances between points from the same cluster will be at most $d + 2\sqrt{d\gamma}+2\gamma$ and all distances between points from different clusters will be at least $d + r^2 - 2\sqrt{d\gamma}$.
According to lemma~\ref{lemma:massart}, this will happen with probability at least $(1 - \exp(-\gamma))^{2n^2 - n}$. 
Denoting this probability as $1 - \epsilon$ and solving for $\gamma$, we obtain
\begin{equation}
    \gamma \le -\log(1-\sqrt[(2n^2 - n)]{1-\epsilon}).\label{eq:k}
\end{equation}
Plugging~\eqref{eq:k} into~\eqref{eq:d} we obtain
\begin{equation}
    d = O\left(\frac{r^4}{-\log(1-\sqrt[(2n^2-n)]{1-\epsilon})}\right).\label{eq:do}
\end{equation}

\begin{figure}[h!]
\begin{center}
\includegraphics[width=0.40\textwidth] {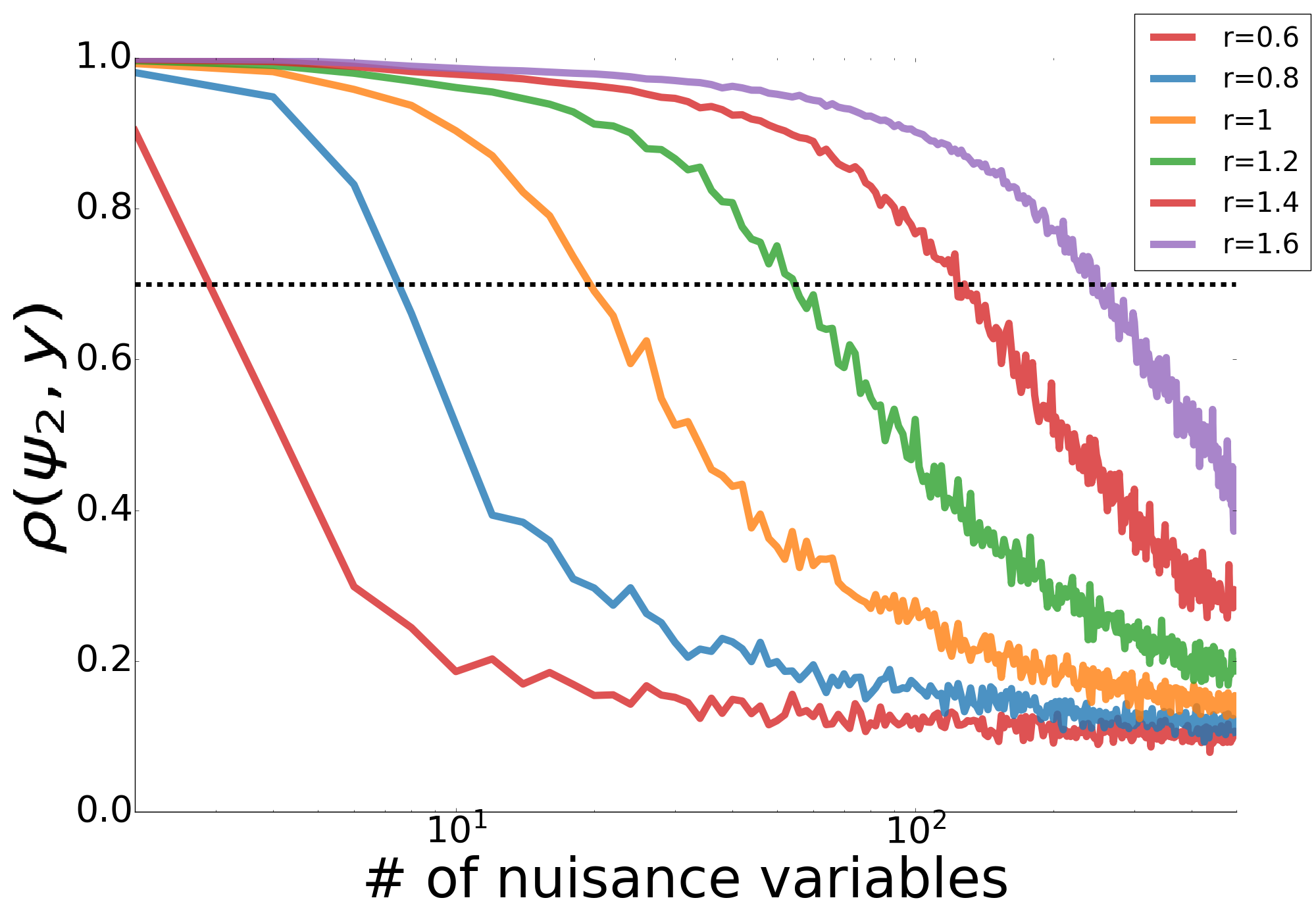}
\includegraphics[width=0.40\textwidth] {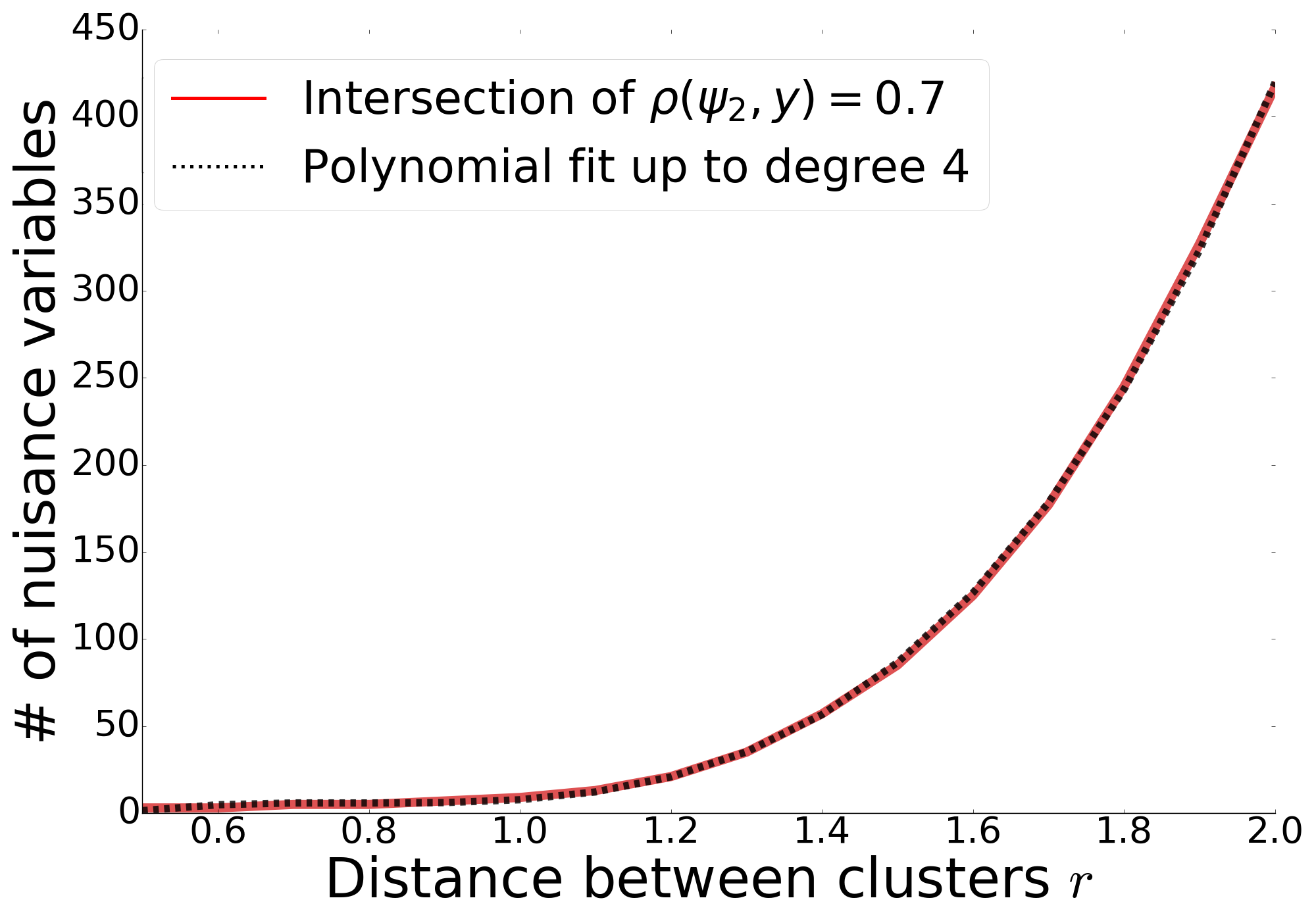}
\end{center}
\caption{Synthetic two cluster datasets. We evaluate the influence of Gaussian nuisance variables on the Laplacian. We generate two clusters using $50$ samples each with distance $r$ apart in $1$-D.
We use $d$ Gaussian nuisance variables and evaluate the leading nontrivial eigenvector $\psi_2$ of the Laplacian. Left: correlation between $\psi_2$ and the true cluster assignments $y$ for different values of $r$. As the number of nuisance variables grows, the eigenvector becomes meaningless. As the distance between clusters decreases, fewer nuisance variables are needed to ``break'' the cluster structure captured by $\psi_2$. Right: by computing the intersection between the damped correlation curves and $0.7$ (shown in the left plot) for different values of $r$, we evaluate the relation between $r$ and the number of nuisance variables $d$ required for breaking the cluster structure. This empirical result supports the analysis presented in \ref{sec:diff} in which we show that $ d = O\left(\frac{r^4}{-\log(1-\sqrt[(2n^2-1)]{1-\epsilon})}\right)$. For convenience, we added a polynomial fit up to degree $4$ presented as the black line. } 
\label{fig:chi}
\end{figure}

In particular, for fixed $n$ and $\epsilon$, equation~\eqref{eq:do} implies that the number of nuisance dimensions must be at most on the order of $r^4$ for the clusters to not mix with high probability. 
In addition, increasing the number of data points for a fixed $r$ and $\epsilon$ brings the argument inside the log term arbitrarily close to zero, which implies that the Laplacian for large data is sensitive to the number of nuisance dimensions.
We support these findings via experiments, as shown in Figure~\ref{fig:chi}.

\section{The Proposed Approach}\label{sec:proposed}
In this section, we present our proposed approach and discuss several of its characteristics.

\subsection{Rational}\label{sec:rational}
In section~\ref{sec:motivation} we demonstrated the problems in computing the Laplacian score using the entire feature set. We concluded that the Laplacian score should ideally be calculated using a Laplacian that is not affected by many nuisance features.
Here we show that this can be tackled by computing the Laplacian score at the CAE concrete layer to achieve a feature selection mechanism that discards both nuisance and correlated ones.

\begin{figure}[h!]
  \centering
    \includegraphics[height=.175\textwidth]{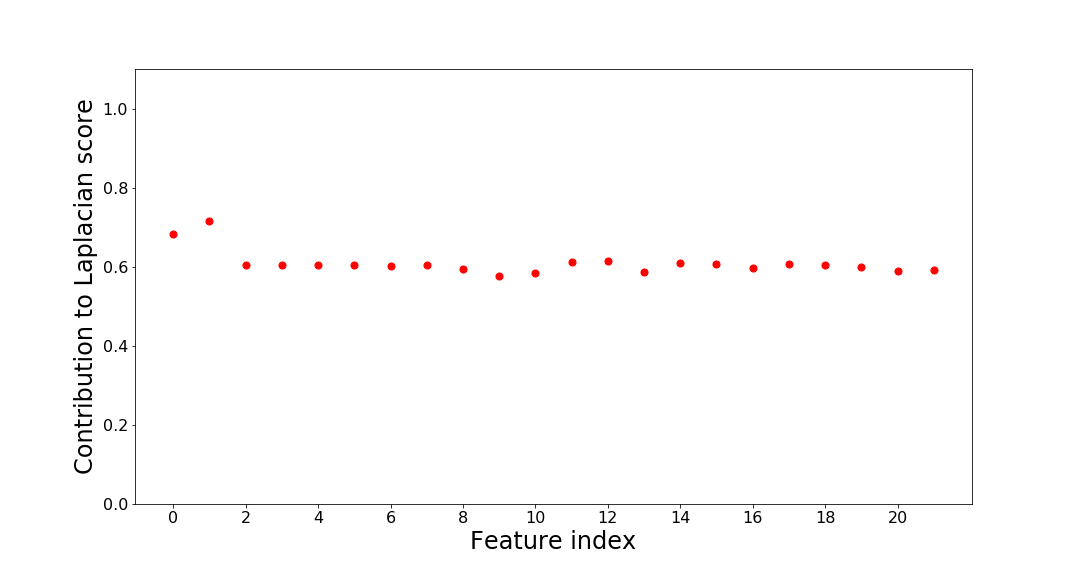}  
    \includegraphics[height=.175\textwidth]{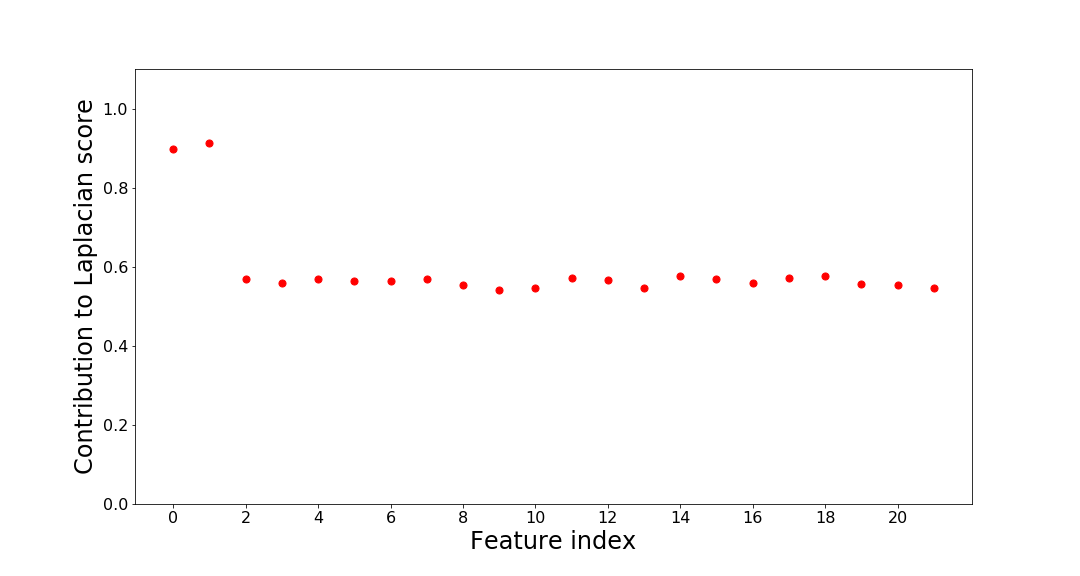}
    \includegraphics[height=.175\textwidth]{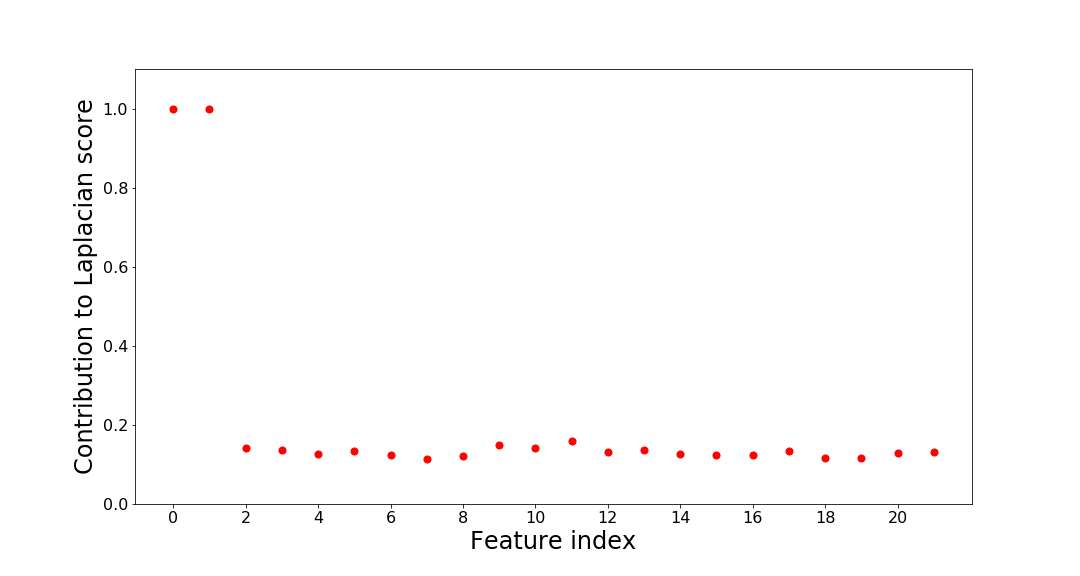}
    \caption{Two moons dataset: contribution $f^TLf$ of each feature $f$ to the Laplacian score. The two leftmost coordinates represent informative (structured) features in all subplots, and the remaining are nuisance features. Left: when the Laplacian is computed using all feature sets, the informative coordinates contribute more to the Laplacian score, which initiates the learning dynamics. Middle: during training, the sampling in the concrete layer gives greater probabilities to sample from the informative coordinates. As a result, the Laplacian score grows. Right: when only informative, structured coordinates are sampled, the Laplacian score is maximized.} 
    \label{fig:ls}
\end{figure}

As explained in Section~\ref{sec:cae}, CAE is an autoencoder model, equipped with a concrete layer in which, at the end of the training, each concrete unit simulates sampling from a categorical distribution with learnable class probabilities.
At the beginning of training, the softmax vector of each concrete unit tends to be flat due to high temperature, as can be seen in equation~\eqref{eq:gs}.
Propagating the data through the concrete layer and using this data representation to compute the Laplacian score creates a corrupted Laplacian, in which the entire feature set, including nuisance features, is taken into account. 
While this is undesirable, as explained above, the contribution of informative features to the Laplacian score often tends to be slightly higher than the contribution of nuisance features, as is noticeable in the left panel of figure~\ref{fig:ls}. 
We utilize this fact to create a learning dynamic that promotes the selection of informative (i.e., not nuisance) features by adding the Laplacian score to the CAE objective function.
Doing so encourages the sampling probabilities of informative features to grow, and this dynamic strengthens during training, as can be seen in the middle and left panels of figure~\ref{fig:ls}.
As is the case for CAE, at the end of the training, the temperature is low, which results in the concrete softmax vectors being approximately one-hot, which effectively simulates a feature selection mechanism.
Hence, by computing the Laplacian score at the CAE concrete layer and adding it to the CAE objective function, one obtains a feature selection mechanism biased towards selecting (i) informative features, which (ii) suffice for the approximate reconstruction of the complete feature set. 

\subsection{LS-CAE}\label{sec:uufs}

In this manuscript, the approach we propose, termed LS-CAE for Laplacian Score-regularized CAE, is an extension of CAE, essentially by adding a Laplacian score term to its objective function during training, where the Laplacian is computed at the concrete layer.

Specifically, the proposed approach inherits from CAE the autoencoder framework and the concrete layer and the reconstruction loss as a feature selection mechanism that promotes selection of a sparse subset of representative features which capture much of the information of the complete feature set.
As this alone does not encourage the discarding of nuisance features, we augment the CAE objective function with a Laplacian score term.
However, armed with the insight that the Laplacian should be computed on the selected features rather than on the complete feature set, we calculate the Laplacian at the concrete layer.

Experimentally, we have noticed that the two objective losses might be of very different magnitudes at different times during training, resulting in a single term dominating the training dynamics. 
To avoid this, we utilize a balancing mechanism that ensures that the two objective components are of similar magnitude and can both affect the training dynamic and the selected features.

More formally, let X be a $m \times d$ minibatch of training data. Denote by $\hat{X}$ the autoencoder output, and denote the output of the concrete layer by $C = C(X)$. Our proposed objective function is therefore
\begin{equation}
    L(X) = \frac{\|X - \hat{X}\|_2^2} {\sg\left( \|X - \hat{X}\|_2^2\right)} - \frac{\trace[C^TL_\text{diff}(C)C]}{\sg\left(\trace[C^TL_\text{diff}(C)C]\right)}, \label{eq:ours}
\end{equation}
where $L_\text{diff}(C)$ is the diffusion Laplacian $D^{-1}W$, computed on the concrete layer representation of the data, and $\sg$ is the Stop Gradient operator, which acts as an identity at forward and has zero partial derivatives.

The balancing mechanism, whose magnitude inversely weights each loss component, removes the need to use a tunable hyperparameter to balance between them and ensures both terms are taken into account in selecting features. 
In addition, we have empirically observed that this results in a more stable training dynamic, comparing to~\citep{lindenbaum2020let}, where the Laplacian score term alone encourages the selection of all features. In contrast, the regularization term encourages the opposite goal, as the opposition forces may result in instability of the training process and increased sensitivity to hyper-parameter tuning.

\subsection{Penalizing Redundant Selection of Same Feature}
The concrete layer mechanism allows scenarios where two or more concrete units select the same input feature. 
While this is wasteful from a reconstruction perspective, the Laplacian score term might benefit from it.
To avoid this, we add a regularization term, penalizing the selection of a feature more than once.
This regularization term is computed as follows
\begin{equation}
    \reg = M\max\{0, m - 1 \},
\end{equation}
where $M$ is a large constant, $m$ is the maximal sum of weights of any features by concrete units, i.e.,
\begin{equation}
    m := \max_{j=1,\ldots d}\sum_{i=1}^k Z_{ij},
\end{equation}
and $Z$ is the $k \times d$-sized matrix of concrete layer probabilities.

\subsection{Temperature Annealing}

As the difference in the contributions to the Laplacian score between nuisance and informative features might be small at the beginning of training when the Laplacian is corrupted (for example, in the left panel of Figure~\ref{fig:ls}), it is beneficial to let this term undergo a warm-up period at the beginning of training before the difference between a nuisance and important features starts to become apparent through the concrete probabilities.

To allow for this warm-up period, we use a linearly decaying temperature annealing schedule, rather than the exponential schedule initially used in the official code of~\cite{balin2019concrete}\footnote{\url{https://github.com/mfbalin/Concrete-Autoencoders}}. 
Effectively, the slower temperature annealing schedule enables various subsets of selected features to be evaluated during training before the probabilities settle on sampling from the desired features. Experimentally we noticed that without changing the annealing schedule, in the absence of the ``warm-up period'', the Laplacian score term was sometimes unable to avoid the selection of nuisance features.


\section{Experimental Results}\label{sec:experiments}

This section provides experimental results on simulated and real-world datasets, demonstrating the proposed approach's efficacy compared to other unsupervised feature selection baselines.
We begin with Ablative experiments, justifying the proposed design, and then to real-world data experiments.

\subsection{Ablation Study}

Our proposed objective function~\eqref{eq:ours} contain a reconstruction term and a Laplacian score term. In this section, we design two simulated experiments. We show that the proposed objective yields a better selection of features compared to methods containing one of the objective terms, but not both. Specifically, we compare our approach (LS-CAE), concrete autoencoder (CAE), which utilizes only the reconstruction term, and a model using only the Laplacian score objective (LS). All models shared an identical architecture and training hyperparameter configuration.

\subsubsection{Simulated Data}
In this experiment, we construct the dataset as follows:
The dataset contains $n=1200$ and $2d+4$ input features, where 2 of the features are the original two moons features, as in the top left panel of Figure~\ref{fig:twomoons}. We then add another noisy copy of the two original features and two copies of $d$ nuisance features, obtained by sampling a multivariate $d$-dimensional multivariate Gaussian, with zero mean and covariance $C$ such that $C_{ij} = (-0.25)^{|i-j|}$.
All models were trained with two concrete units. We measured the proportion of times where the two selected input features were the two original two moon coordinates (from either of the two copies).

The dataset was constructed this way to demonstrate that in the presence of correlated features: (i) the reconstruction term alone might favor features which allow for low reconstruction error, despite being high-frequency (i.e., irrelevant to the cluster structure) and (ii) the Laplacian score term might favor two copies of the same original input feature and ignore other low-frequency features. 
The dataset was $z$-transformed before training the model so that all features had zero mean and unit standard deviation.

\begin{figure}[h!]
  \centering
    \includegraphics[height=.375\textwidth]{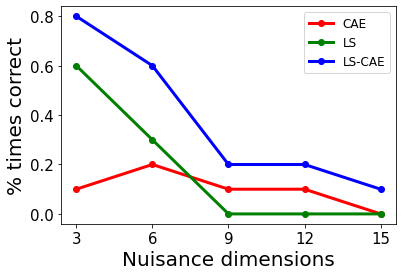}  
    \caption{Ablation study: augmented two moons data. } 
    \label{fig:ablation1}
\end{figure}

We trained the model for $d=3, 6, 12, 15$, with ten repetitions per setting (where repetitions differ in the sampling of the dataset and the initialization of the model).
Figure~\ref{fig:ablation1} shows the results of this experiment. As can be seen, having both objective terms consistently (overall values of $d$) yields the selection of better features than just one of the objectives but not the other objective in this case.

\subsubsection{MNIST}
\begin{figure}[h!]
  \centering
    \includegraphics[width=.25\textwidth]{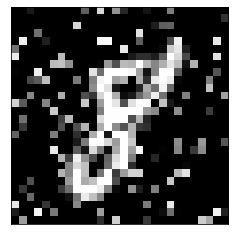}  
    \includegraphics[width=.25\textwidth]{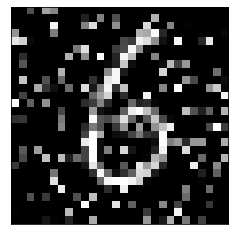}
    \includegraphics[width=.25\textwidth]{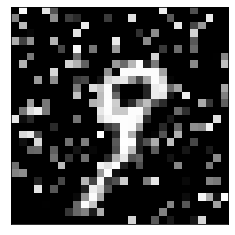}
    \caption{Examples from the noisy mnist dataset. } 
    \label{fig:mnist_exmples}
\end{figure}

In this experiment we create a noisy version of the MNIST handwritten digit dataset, via replacing $x_i$ with $\min\{0, \max \{255, x_i  + np \odot m_i\}\}$, where $np$ is a $28\times 28$ noise pattern sampled a i.i.d uniform distribution over $\{0,1,\ldots,255\}$, $m_i$ is a $28\times 28$ i.i.d Bernoulli$(0.2)$ mask and $\odot$ denotes element-wise product. 
As the noise pattern, $np$ is common to all images; the data contains correlated high-frequency features, which might lead to the sub-optimal selection of features using the reconstruction term alone. In addition, since adjacent pixels are typically highly correlated in the MNIST datasets, the Laplacian score term might favor selecting such pixels while ignoring other other features that are important to identify the image type. 
Examples of noisy images are shown in Figure~\ref{fig:mnist_exmples}

\begin{figure}[h!]
  \centering
    \includegraphics[width=.65\textwidth]{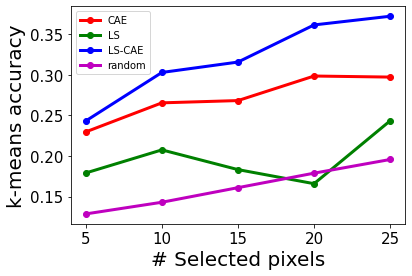}  
    \caption{Results of the noisy MNIST experiment. } 
    \label{fig:mnist_results}
\end{figure}

We trained the models to select $5, 10, 15, 20$ and $25$ features. Once the features were selected, we trained a $k$-means with $k=10$ on the training dataset (60,000 examples), using only the selected features, and measured the clustering accuracy on the test dataset (10,000 examples). 
For each number of features, we repeated the above procedure three times. The results are presented in Figure~\ref{fig:mnist_results}, which shows the average clustering accuracy of each of the methods, and also the clustering accuracy obtained when the features are selected randomly.

\begin{figure}[h!]
  \centering
    \includegraphics[width=.25\textwidth]{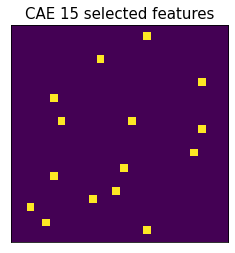}
    \includegraphics[width=.25\textwidth]{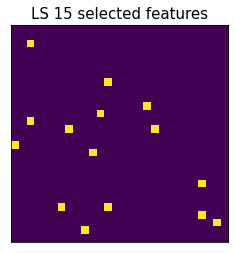}
    \includegraphics[width=.25\textwidth]{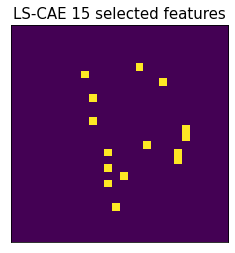} 
    \caption{Noisy MNIST experiment: example of 15 features selected by each of the methods.} 
    \label{fig:mnist_features}
\end{figure}

As evident from this plot, the proposed approach consistently leads to higher clustering accuracies (compared to the other baselines). This suggests that it can identify a better subset of features that carry information of the main underlying structure in the data.
Figure~\ref{fig:mnist_features} shows examples of the features selected by each of the methods. 
As can be seen, the added noise makes CAE select pixels scattered over a large portion of the image, many of which are not indicative of the digit type. About half of the features selected by the Laplacian score term are near the boundary of the image, as they are low-frequency but irrelevant for identifying the digit type. On the other hand, LS-CAE seems to select pixels concentrated in areas of the images relevant for determining the cluster component the image belongs to.

\subsection{Real world datasets}
To demonstrate the advantage of using LS-CAE on real-world data, we now turn our attention to nine publicly available feature selection benchmark datasets \footnote{https://jundongl.github.io/scikit-feature/datasets.html}.
Table~\ref{tab:stats} summarizes the properties of the datasets used for these experiments. The datasets vary in sample size, from as few as 56 examples to 21,332. In addition, they contain a large number of features, often higher than the sample size. 
On such datasets, the quality of the set of selected features can dramatically improve the performance of downstream tasks, such as clustering, as will be demonstrated next. 
The $k$-means clustering accuracy on each dataset using all features (i.e., without any feature selection) is indicated in Table~\ref{tab:stats} as well.

\begin{table*}[h!]
  \centering

  \begin{adjustbox}{width=.6 \columnwidth,center}
  \small{
    \begin{tabular}{lllll }
    \hline
    Datasets & Dim    & Samples  & Classes  & Accuracy using All features \\
    \hline
    RCV1 & 24408 & 21332 & 2 & 50.0 \\
    GISETTE & 4955 & 6000 & 2 & 74.4\\
    PIX10 &  10000 & 100 & 10& 74.3\\
    COIL20 & 1024 & 1444 & 20 & 53.6\\
    Yale  &  1024 & 165 & 15 & 38.3\\
    TOX-171 & 5748 & 171 & 4 & 41.5\\
    ALLAML &  7192 & 72 & 2 & 67.3\\
    PROSTATE & 5966 & 102 & 2 & 58.1\\
    FAN &  25683 & 56 & 8 & 37.5\\
    POLLEN & 21810 & 301 & 4 & 54.9\\    
    \end{tabular}%
    }
\end{adjustbox}

  \caption{Properties of each of the real-world benchmark datasets used in the experiments.} 
   \label{tab:stats}

\end{table*}%

We compare to several strong baselines, such as Laplacian Score \cite{he2006laplacian} (LS), Multi-Cluster Feature Selection \cite{cai2010unsupervised} (MCFS), Local Learning based Clustering (LLCFS) \cite{zeng2010feature}, Nonnegative Discriminative Feature Selection (NDFS) \cite{li2012unsupervised}, Multi-Subspace 
Randomization and Collaboration (SRCFS) \cite{huang2019unsupervised}, and Concrete Auto-encoders (CAE) \cite{balin2019concrete}. 
Each model is tuned to select the best $50,100,150,200,250,$ or $300$ features. Then, we apply $k$-means $20$ times on the selected features and compute the average clustering accuracy. For each method, we report the highest (average) clustering accuracy along with the number of selected features.

\begin{table*}[h!]
  \centering

  \begin{adjustbox}{width=.9 \columnwidth,center}
  \small{
    \begin{tabular}{llllllllll }
    \hline
    Datasets & LS    & MCFS  & NDFS  & LLCFS & SRCFS  & CAE    & LS-CAE\\
    \hline
        RCV1 & 54.9 (300) & 50.1 (150) & 55.1 (150) & 55.0 (300) & 53.7 (300) &  54.9 (300) & {\bf 83.7 (300)}\\
     GISETTE & 75.8 (50) & 56.5 (50) & 69.3 (250) &  72.5 (50) & 68.5 (50) & 77.3 (250)     & {\bf 80.7 (50)}\\
    PIX10 & 76.6 (150) & 75.9 (200) & 76.7 (200)&  69.1 (300) & 75.9 (100) & {{94.1 (250)} }     & {\bf{94.5 (250)} }\\
    COIL20 & 60.0 (300) & 59.7 (250) & 60.1 (300) & 48.1 (300) & 59.9 (300) & {\bf{65.6 (200)}}  &   61.8 (300)\\
    Yale  & 42.7 (300) & 41.7 (300)  & 42.5 (300) & 42.6 (300) & 46.3 (250) & 45.4 (250)  & {\bf{48.0 (200)}}\\
    TOX-171 & 47.5 (200) & 42.5 (100) & 46.1 (100) & 46.7 (250) & 45.8 (150) &  47.7 (100)   & {\bf{48.3 (100)}}\\
    ALLAML & 73.2 (150) & 72.9 (250) & 72.2 (100) & {\bf{77.8 (50)}} & 67.7 (250) & 73.5 (250) & 76.5 (150)\\
    PROSTATE & 58.6 (300) & 57.3 (300) & 58.3 (100) & 57.8 (50) &60.6 (50) & 56.9 (250)    & \bf {71.4 (50)}\\
        FAN & 42.9 (150) & 45.5 (150) & 48.8 (100) &  29.0 (50) & 29.0 (100) & 35.2 (300)   & {\bf 51.7 (100) }\\
  POLLEN & 46.9 (150) &{\bf{ 66.5 (300) }} & 48.9 (50) &  35.0 (100) & 34.9 (300) & 58.0 (250)  & 65.8 (100) \\    
    \hline
        Mean rank & 4.0 & 6.0 & 5.0 & 5.0 & 6.0 & 2.0&  \bf{1.0}&  \\
        Median rank & 3.67 & 5.89 & 4.33 & 4.67&  5.22 & 2.89&  \bf{1.33} \\

    \hline
    \end{tabular}%
    }
\end{adjustbox}

  \caption{Average clustering accuracy on several benchmark datasets. Clustering is performed by applying $k$-means to the features selected by the different methods. The number of selected features is shown in parenthesis.} 
   \label{tab:results}

\end{table*}%
As can be seen, on seven of the ten benchmark datasets, LS-CAE achieves the best performance, and the second-best on the remaining three, with an average rank of 1.
Noticeably, on the RC1 benchmark, our proposed approach outperforms the following best method by 50\%. On the Prostate and FAN dataset, our proposed approach also significantly improves more than 10\% over the following best methods.

\subsection{Technical details}
All the experiments reported in this manuscript were performed using the same decoder architecture, containing two hidden layers, each of 128 LeakyReLU units.
We trained the models for 300 epochs, using a learning rate of 1. for the concrete layer and 0.01 for the decoder.


\section{Conclusion}\label{sec:conclusion}

In this manuscript, we propose an unsupervised approach for feature selection, utilizing concrete layer mechanism and Laplacian score and reconstruction objectives.
We demonstrated both analytically and numerically that for the Laplacian score to be a helpful criterion in the presence of many nuisance features, it is crucial to be computed on the subset of selected features rather than on the complete feature set. 
We showed that the proposed objective function's two components are needed to avoid both high frequency and correlated features and reported state-of-the-art results on several real-world datasets of various sizes.

\bibliography{references}

\end{document}